\def\year{2020}\relax
\documentclass[letterpaper]{article} 
\usepackage{aaai20}  
\usepackage{times}  
\usepackage{helvet} 
\usepackage{courier}  
\usepackage[hyphens]{url}  
\usepackage{graphicx}  
\usepackage{multirow}
\usepackage{amsmath}
\usepackage{graphicx, subfig}
\usepackage{url}

\title{Span Model for Open Information Extraction on Accurate Corpus}
\author{Junlang Zhan\textsuperscript{\rm 1,\rm 2,\rm 3}, Hai Zhao\textsuperscript{\rm 1,\rm 2,\rm 3,}\thanks{Corresponding author. This paper was partially supported by National Key Research and Development Program of China (No. 2017YFB0304100), 
Key Projects of National Natural Science Foundation of China (U1836222 and 61733011).}\\ 
\textsuperscript{\rm 1}Department of Computer Science and Engineering, Shanghai Jiao Tong University\\ 
\textsuperscript{\rm 2} Key Laboratory of Shanghai Education Commission for Intelligent Interaction \\
and Cognitive Engineering, Shanghai Jiao Tong University, Shanghai, China \\
\textsuperscript{\rm 3} MoE Key Lab of Artificial Intelligence, AI Institute, Shanghai Jiao Tong University \\
longmr.zhan@sjtu.edu.cn, zhaohai@cs.sjtu.edu.cn 
}
 
\begin{document}

\maketitle

\begin{abstract}
Open Information Extraction (Open IE) is a challenging task especially due to its brittle data basis. Most of Open IE systems have to be trained on automatically built corpus and evaluated on inaccurate test set. In this work, we first alleviate this difficulty from both sides of training and test sets. For the former, we propose an improved model design to more sufficiently exploit training dataset. For the latter, we present our accurately re-annotated benchmark test set (Re-OIE2016) according to a series of linguistic observation and analysis. Then, we introduce a span model instead of previous adopted sequence labeling formulization for $n$-ary Open IE. Our newly introduced model achieves new state-of-the-art performance on both benchmark evaluation datasets.
\end{abstract}

\section{Introduction}
Open Information Extraction (Open IE) aims to generate a structured representation of information from natural language text in the form of verbs (or verbal phrases) and their arguments. An Open IE system is usually domain-independent and does not rely on a pre-defined ontology schema. An example of Open IE is shown in Table~\ref{simple_example}. Open IE has been widely applied in many downstream tasks such as textual entailment and question answering \cite{Fader2014OpenQA}. Some existing Open IE systems are listed in Table~\ref{OIE_systems}. Most of these systems are built in unsupervised manner except that TextRunner and OLLIE are built in a self-supervised manner.

To perform Open IE in supervised learning and to leverage the advantages of neural network, \cite{Stanovsky2018SupervisedOI} created the first annotated corpus by an automatic translation from Question-Answer Driven Semantic Role Labeling (QA-SRL) annotations \cite{He2015QuestionAnswerDS} and Question-Answer Meaning Representation (QAMR) \cite{Michael2018CrowdsourcingQM}. Based on this corpus, they developed an Open IE system RnnOIE using a bidirectional long short-term memory (BiLSTM) labeler and BIO tagging scheme which built the first supervised learning model for Open IE. \cite{Cui2018NeuralOI} constructed a large but noisy annotated corpus by using OpenIE4 to perform extractions on Wikipedia then kept only tuples with high confidence score. They also built an Open IE system by using a neural sequence to sequence model and copy mechanism to generate extractions. However, this system can only perform binary extractions.

\begin{table}[t!]
\begin{center}
\begin{tabular}{|l|l|}
\hline 
Sentence & \emph{Repeat customers can purchase} \\ 
& \emph{luxury items at reduced prices.}\\ 
\hline 
Extraction & (A0: repeat customers; \textbf{can purchase}; \\
&A1: luxury items; A2: at reduced prices)\\
\hline
\end{tabular}
\end{center}
\caption{An example of Open IE in a sentence. The extraction consists of a predicate (in bold) and a list of arguments, separated by semicolons.}
\label{simple_example}
\end{table}

\begin{table*}[t!]
\begin{center}
\begin{tabular}{|l|c|c|l|}
\hline 
\multicolumn{2}{|c|}{\textbf{Systems}}   & \multirow{2}{*}{\textbf{Base}} &\multirow{2}{*}{\textbf{Main Features}} \\ 
\cline{1-2} 
\textbf{Name} & \textbf{Work} & & \\
\hline 
\multirow{2}*{TextRunner} & \multirow{2}*{\cite{Yates2007TextRunnerOI}} & \multirow{2}*{-} & The first Open IE system using\\
&&&self-supervised learning approach. \\
\hline

ReVerb   & \cite{Fader2011IdentifyingRF} & - & Leveraging POS tag patterns. \\
\hline 

 OLLIE &  \cite{Mausam2012OpenLL} & Reverb & Learning extraction patterns. \\
\hline

ClausIE & \cite{Corro2013ClausIECO} & - & Decomposing a sentence into clauses. \\ \hline

Stanford& \cite{Angeli2015LeveragingLS}& -  & Splitting a sentence into utterances.\\
\hline

\multirow{2}*{OpenIE4} & \multirow{2}*{\cite{Mausam2016OpenIE}} & \multirow{2}*{OLLIE} & Adding  RELNOUN \cite{Pal2016DemonymsAC}\\
&&&, SRLIE \cite{Christensen2011AnAO}.\\
\hline

\multirow{2}*{OpenIE5} & \multirow{2}*{-} & \multirow{2}*{OpenIE4} & Adding CALMIE \cite{Saha2018OpenIE} \\
&&&, BONIE \cite{Saha2017BootstrappingFN}.\\
\hline

\multirow{2}*{PropS}  & \multirow{2}*{\cite{Stanovsky2016GettingMO} }&\multirow{2}*{-}&Using conversion rules from \\
&&&dependency trees.\\
\hline

MinIE  & \cite{Gashteovski2017MinIEMF} &ClausIE& Mitigating overspecific extractions.\\
\hline

\multirow{2}*{Graphene}  &\multirow{2}*{\cite{Cetto2018GrapheneAC}}&\multirow{2}*{-}&Parsing a sentence into core facts \\
&&&and accompanying contexts.\\
\hline

RnnOIE &\cite{Stanovsky2018SupervisedOI}&-&BiLSTM and BIO tagging.\\
\hline

Seq2seq OIE & \cite{Cui2018NeuralOI}&-& Seq2seq model and copy mechanism.\\
\hline

\end{tabular}
\end{center}
\caption{A summary of existing Open IE systems}
\label{OIE_systems}
\end{table*}

In this work, we develop an Open IE system by adapting a modified span selection model which has been applied in Semantic Role Labeling (SRL)\cite{Ouchi2018ASS,he-etal-2018-syntax,Li2019DependencyOS,Zhao2013IntegrativeSD,zhao-etal-2009-semantic,li-etal-2018-unified,cai2018full}, coreference resolution\cite{shou-zhao-2012-system,zhang-etal-2012-chinese} and syntactic parsing \cite{zhang-etal-2016-probabilistic,ma-zhao-2012-fourth,Li2020GlobalGreedy,zhou-zhao-2019-head,zhao-etal-2009-cross,li-etal-2018-seq2seq}. The advantage of a span model is that span level features can be sufficiently exploited which is hard to perform in token based sequence labeling models. As far as we know, this is the first attempt that applies span model on Open IE. 

As previous Open IE systems were mostly built on a loose data basis, we thus intend to strengthen such a basis from both sides of training and test sets. We constructed a large training corpus following the method of \cite{Cui2018NeuralOI}. The differences of our construction method and the method of \cite{Cui2018NeuralOI} are two-fold: First, our corpus is constructed for $n$-ary extraction instead of binary extraction. Second, previous methods only keep extractions with high confidence score for training. However, the confidence scores estimated by OpenIE4 are not so reliable and we have found that some extractions with low confidence score are also useful. We thus propose a method designed to leverage such extractions with low confidence scores.

At the stage of evaluation, the benchmark constructed by \cite{Stanovsky2016CreatingAL} is widely used as the test set for Open IE performance evaluation. However, since this corpus was automatically constructed from QA-SRL, we observe that there are still incorrect extractions. To reduce the noise and obtain a more accurate evaluation for Open IE systems, we manually relabel this corpus. Experiments show that our system outperforms previous Open IE systems in both old benchmark and our relabeled benchmark. 

\section{Model}
Our system, named SpanOIE, consists of two parts. The first part is the predicate module to find predicate spans in a sentence. The second part is the argument module which takes a sentence and a predicate span as input and outputs argument spans for this predicate. 

\subsection{Problem Definition}
We formulize the open information extraction as a span selection task. A span (\textit{i, j}) is a piece of a sentence beginning with the word \textit{i} and ending with the word \textit{j}. The goal is to select the correct span for each Open IE role label. Given a sentence $W=w_1,...,w_n$, our model predicts a set of predicate-argument relations $Y \subseteq  S \times P \times L$, where $S=\{(w_i,...,w_j)|1\leq i \leq j \leq n\}$ is the set of all the spans in $W$, $P$ is the set of predicate spans which is a subset of $S$ and $L=\{A0, A1, A2, A3\}$ is the label set of Open IE\footnote{According to \cite{Stanovsky2018SupervisedOI}, this labeled set covers more than 96\% sentences.}. For each label $l$, our model selects a span $(i,j)$ with the highest score:
\begin{equation}
    \mathop{\arg\max}_{(i',j')\in S} SCORE_l (i',j'), l \in L
\end{equation}
where
\begin{equation}
    \begin{split}
      SCORE_l (i,j)= P_{\theta}(i,j|l) \\ = \frac{{\rm exp}(\phi_\theta (i,j,l))}{\sum_{(i',j') \in S} {\rm exp}(\phi_\theta (i',j',l))}  
    \end{split}
\end{equation}
and $\phi_\theta$ is a trainable scoring function with parameters $\theta$. To train the parameters $\theta$, in the training set, for each sample $X$ and the gold structure $Y^*$, we minimize the cross-entropy loss:
\begin{equation}\label{loss}
    l_\theta(X,Y^*)=\sum_{(i,j,l)\in Y^*} -{\rm log} P_{\theta}(i,j|l)
\end{equation}

Note that some labels may not appear in the given sentence. In this case, we define the predicate span as NULL span and train a model to select the NULL span.

\subsection{Spans Candidates Selection}
The main drawback of the span model is that too many spans generated from a sentence will cause too high computational cost and hurt the performance of the model. The number of all possible spans for a sentence with size $T$ is $\frac{T(T+1)}{2}$, which makes the number of the spans grow quickly with the length of the sentence. To reduce the number of spans, we propose three constraints for span candidates:
\begin{itemize}
\item {\textbf{Maximum length constraint}}: During training, we only keep the argument spans whose size is less than 10 words and the predicate spans whose size is less than 5 words. At inference stage, we remove this constraint.
\item {\textbf{No overlapping constraint}}:  We keep only the spans which do not overlap with the given predicate span.
\item {\textbf{Syntactic constraint}}: This constraint is only for spans whose length is bigger than 1. We keep only the spans which satisfy that a word is either the syntactic parent of another word in the same span, or the parent of this word is in the same span. For the example shown in Figure~\ref{model}, the span [\emph{luxury items at}] violates this syntactic constraint because the word \emph{at} is not the parent of a word in the same span and its parent, the word \emph{purchase} is not in this span.
\end{itemize}

\subsection{Model Architecture}

\begin{figure*}[ht]
\begin{center}
\includegraphics[scale=0.5]{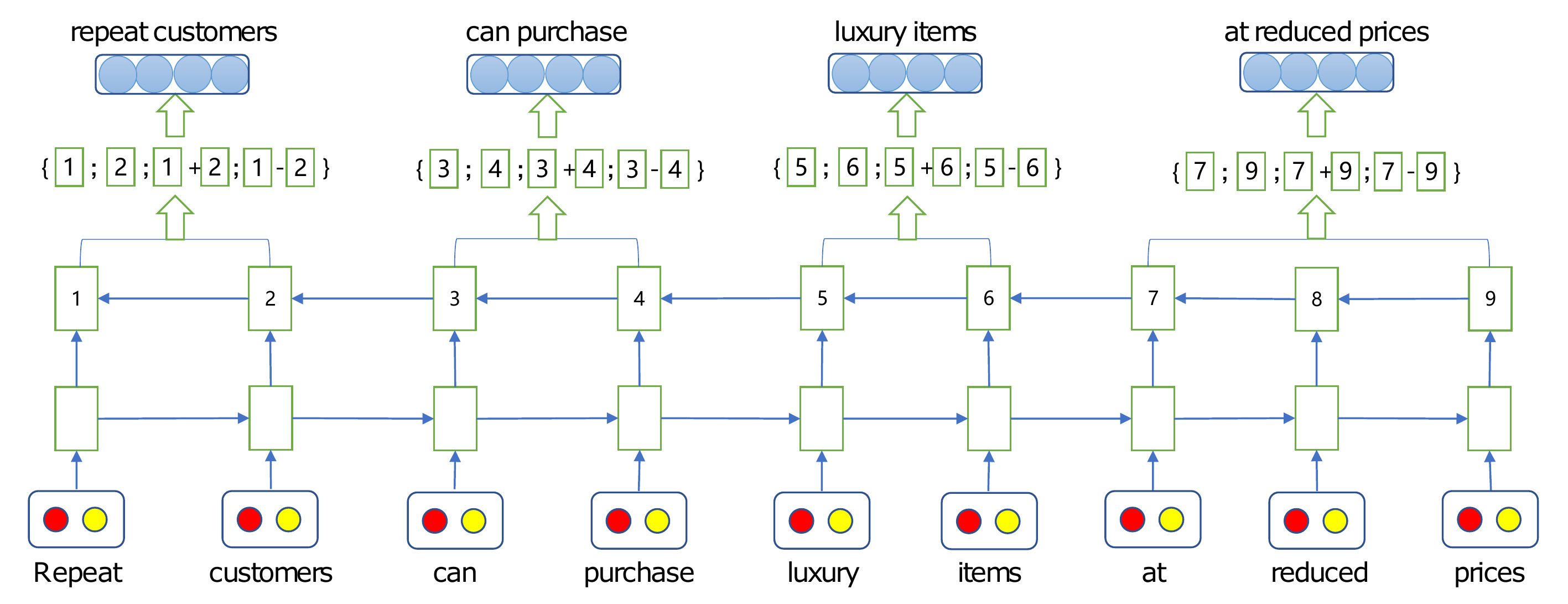}
\caption{An overview of the span model}
\label{model}
\end{center}
\end{figure*}

As our model works as two-stage pipeline, there are two modules for predicate and argument modeling which are implemented in the same way as a BiLSTM network shown in Figure~\ref{model}. 

Given an input sample which contains a sentence $S$ and a predicate span $P$\footnote{The argument span model will takes the known predicate span as input, while for the predicate span model, it will only take the sentence as input.}, we extract a feature vector $x_i$ for every word $w_i \in S $:
\begin{equation}
    \begin{split}
        x_i=emb(w_i) \oplus emb(pos(w_i))
        \\ \oplus  emb(p(w_i)) \oplus emb(dp(w_i))
    \end{split}
\end{equation}
where $\oplus$ denotes concatenation, $emb(w_i)$ is the word embedding, $emb(pos(w_i))$ is the POS tag embedding, $emb(p(w_i))$ is the embedding of a binary value which indicates whether $w_i$ belongs to predicate span and $emb(dp(w_i))$ is the embedding of dependency relation between $w_i$ and its syntactic parent.

The feature vector $x_i$ is then fed into a BiLSTM network which computes contextualized output features:
\begin{equation}
    \overrightarrow{h_i},\overleftarrow{h_i} = BiLSTM(x_i)
\end{equation}
The forward and the backward output of the BiLSTM are then concatenated together:
\begin{equation}
    h_i = \overrightarrow{h_i}\oplus\overleftarrow{h_i}
\end{equation}
We take span features as follows:
\begin{equation}\label{span_feature}
    \begin{split}
        f_{span}(s_{i:j}) = h_i \oplus h_j \oplus h_i + h_j\\
         \oplus hi - h_j 
    \end{split}
\end{equation}
where $s_{i:j}$ is the span which starts from $w_i$ and ends at $w_j$. $h_i$ and $h_j$ are used as a part of span features in the works of \cite{He2018JointlyPP}. $h_i+h_j$ and $h_i-h_j$ are used as span features in \cite{Ouchi2018ASS}. The span features are then fed into a feed forward network and later a Softmax layer to obtain the scores of different labels for each span. For the predicate model, we set binary labels to indicate if the span is a predicate or not. For the argument model, the predicted labels directly correspond to the argument labels. 

\subsection{Inference}
 At the inference stage, several effective decoding methods have been proposed for the span selection model such as structural constraint inference by using integer linear programming  and dynamic programming. In this work, we use  span-consistent greedy search \cite{Ouchi2018ASS}. 

\textbf{Predicate Inference} A predicate span that is completely included in another predicate span in the same sentence will be dropped. For example, given a sentence [\emph{James wants to sell his company}], two spans [\emph{wants to sell}] and [\emph{to sell}] are both selected as predicates, we keep only the former one. This strategy is reasonable since [\emph{James; to sell; his company}] is not an exact fact that can be derived from the original sentence.

\textbf{Argument Inference} Given a sentence and a predicate span, our argument model first scores every possible span for each label in the sentence. Then all the tuples ($span$, $label$, $score$) are sorted according to their scores and top scored tuples are selected. If a tuple with a specific label was not selected before, then the tuple will be selected. The selection ends until all the labels have their corresponding span. If the corresponding span of a label is the NULL span, this label will be dropped in the extraction.

\subsection{Confidence Score}
Like other Open IE systems, our model also provides a confidence score for every extraction. The confidence score $CS$ of an extraction is defined as follows:
\begin{equation}
        CS = score_{pred} + \sum_{i=1}^{label\_number}score_{i,span_i}
\end{equation}
where $score_{pred}$ is the score for the predicate given by the output of Softmax layer of the predicate model, $score_{i,span_i}$ is the score of selected span with label $i$ given by the span model.

\section{Usage of Training Corpus}
We use a part of raw corpus that is preprocessed by \cite{Cui2018NeuralOI}\footnote{\url{https://1drv.ms/u/s!ApPZxTWwibImHl49ZBwxOU0ktHv}}. OpenIE4\footnote{\url{https://github.com/allenai/openie-standalone}} is used to do the $n$-ary extraction. Although there exists OpenIE5\footnote{\url{https://github.com/dair-iitd/OpenIE-standalone}}, which is an advanced version of OpenIE4, we still use the latter due to too high computational cost \footnote{The processing time is much longer in OpenIE5 than in OpenIE4} required by the former. Actually, The extractions of OpenIE4 and OpenIE5 are the same for most cases and the major improvements from OpenIE5 are about numerical information extraction and splitting conjunctive sentences, which are not a major concern for our preprocessing. 

\cite{Cui2018NeuralOI} adopt a preprocessing to keep only the extractions with the confidence score greater than 0.9 for training. However, we observe that the confidence scores provided by OpenIE4 are not all reliable. For example, in the sentence [\emph{he makes a state visit}], The extraction by OpenIE4 is [\emph{he; makes; a state visit}]. This extraction looks quite correct while its confidence score is only 0.388 out of 1. We further observe that all the extractions that contain \emph{he, she, they, it} are given low scores. Since Open IE surely belongs to the task of information extraction which is supposed to care about non-trivial constituents, and pronoun seems less interesting to the task. However, we argue that extraction with pronoun may sometimes be quite useful. For example, if the sentence [\emph{he makes a state visit}] is in the article about Barack Obama, then $he$ is referred to the non-trivial named entity Barack Obama. Ignoring the $he$ will cause a loss of critical clue. The pronouns with low scores are just one of many scoring problems caused by OpenIE4 according to our empirical observation. As \cite{Cui2018NeuralOI} takes an absolute threshold 0.9 to filter out all extractions below the threshold, we may lose some useful training data. Thus instead of setting a better threshold, we propose to make use of all the extractions from OpenIE4 by adapting our model loss function in Equation (\ref{loss}) to accommodate the confidence scores $CS_4$ given by OpenIE4, 
\begin{equation}
        l_\theta(X,Y^*)=CS_4(Y^*)\times 
        \sum_{(i,j,l)\in Y^*} -log P_{\theta}(i,j|l)
\end{equation}
In detail, we leverage all the extractions obtained by OpenIE4 for training but for every extraction, we multiply the cross entropy by its confidence score according to OpenIE4. In this way, open to all extractions, our span model may automatically determine the selection of spans, for either those correct extractions but with low confidence scores or those truly bad extractions with high confidence scores. 

In this training corpus, there are 1,109,411 sentences and 2,175,294 extractions. In addition, we annotated this corpus with POS tag and dependency parsing information by using spaCy \footnote{\url{https://spacy.io}}. The positions of the first word and the last word for every span are provided.

\section{Test Corpus: Relabeled OIE2016}
OIE2016 is a broadly adopted evaluation benchmark corpus for open IE, which was used as test set created by \cite{Stanovsky2016CreatingAL}\footnote{\url{https://github.com/gabrielStanovsky/oie-benchmark}}. This corpus which consists of 600 sentences \footnote{Originally, there are 3200 sentences in OIE2016. In this paper, we only consider a subset of this corpus. This subset is also used as test data in \cite{Stanovsky2018SupervisedOI}.} is the processing results of QA-SRL \cite{He2015QuestionAnswerDS}.

However, we have observed some problematic annotations and extractions in OIE2016:
\begin{enumerate}
\item OIE2016 considers all the verbal adjectives as predicates, which causes overspecification of extractions. For the sentence in Figure~\ref{model}, OIE2016 gives an extraction [\emph{reduced; prices}]. In the construction of QA-SRL corpus, annotators were asked to find all possible arguments for every verb, which causes such an extraction overspecification. 
\item OIE2016 considers appositive as A0 while ignoring the appositive relations. For the sentence [\emph{Now Mr. Broberg , a lawyer , claims he 'd play for free}], OIE2016 gives two extractions, [\emph{Mr. Broberg; claims; that he 'd play for free}] and [\emph{a lawyer; claims; that he 'd play for free}], in which the latter is less meaningful. In addition, a meaningful extraction [\emph{Mr. Broberg; be; a lawyer}] is missing. 
\item There are numbers of incorrect extractions due to the wrong translation from QA-SRL. For the sentence  [\emph{A cafeteria is located on the sixth floor , a chapel on the 14th floor , and a study hall on the 15th floor}] , OIE2016 gives two incorrect extractions, [\emph{a study hall; located; on the sixth floor}] and [\emph{a study hall; located; on the 14th floor}]. 
\end{enumerate}

To fix all the above problems, we manually re-annotated the entire OIE2016 corpus for a more accurate evaluation benchmark. The re-annotated OIE2016, named Re-OIE2016, has solved the problem of extraction overspecification, and provides extractions for appositive relation. Besides the predicates and arguments, we also annotated context information in Re-OIE2016. However, since some Open IE systems (including ours) do not perform the context extraction as it is hard to judge the correct boundary for context, this information does not serve as a part of evaluation in this work and will be exploited in future research.

\section{Experiments\footnote{Our code and annotated corpus are publicly available in \url{https://github.com/zhanjunlang/Span_OIE}}}
\subsection{Experiment Settings}
In the experiments, we use pre-trained GloVe \footnote{\url{https://nlp.stanford.edu/projects/glove/}} as as word embedding. Other experiment settings are shown in Table~\ref{parameters}.

\begin{figure*}[h]
\begin{minipage}[t]{0.48\linewidth}
\begin{center}
\includegraphics[scale=0.55]{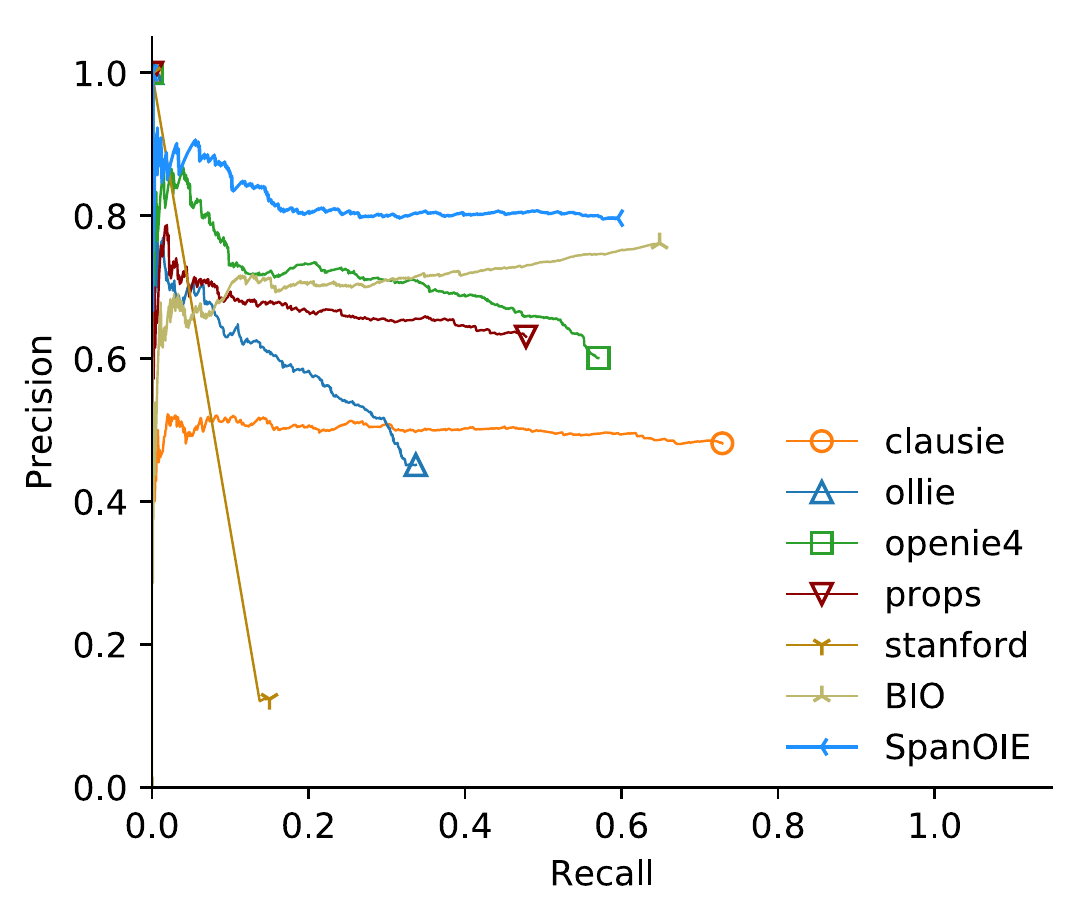}
\end{center}
\caption{The P-R curve of different Open IE systems on OIE2016}
\label{results_old}
\end{minipage}%
\hfill
\begin{minipage}[t]{0.48\linewidth}
\begin{center}
\includegraphics[scale=0.55]{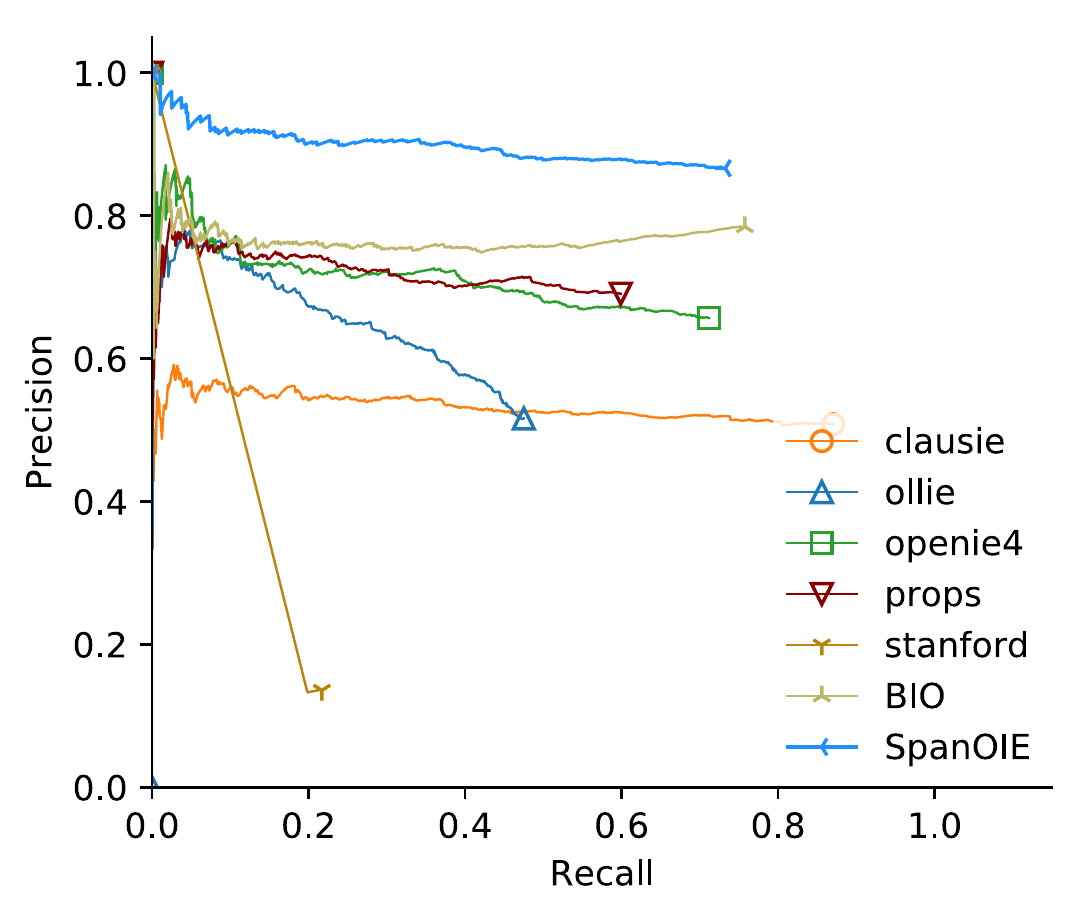}
\end{center}
\caption{The P-R curve of different Open IE systems on Re-OIE2016}
\label{results_new}
\end{minipage}
\end{figure*}

\begin{table}[t!]
\begin{center}
\begin{tabular}{|l|l|}
\hline 
\textbf{Model Parameters} & \\
\hline 
Word embedding size & 100 \\ 
POS tag embedding size & 10\\
Dependency label size & 20 \\
Hidden size of BiLSTM & 200 \\
Number of layers of BiLSTM & 2 \\
\hline 
\textbf{Training Parameters} & \\
\hline 
Batch size & 20\\
Training steps & 55470\\
Optimizer & Adam \\
Learning rate & 0.01 \\
Dropout & 0.3 \\
Decay rate for every 100 steps & 0.005 \\
\hline
\end{tabular}
\end{center}
\caption{Experiment settings}
\label{parameters}
\end{table}

\subsection{Open IE Systems}
We will evaluate the performance of different Open IE systems on both OIE2016 and Re-OIE2016. These systems include ClausIE, OLLIE, OpenIE4, Props, Stanford OIE and our system SpanOIE. To demonstrate the advantage of the span model compared to the sequence labeling model, we also realized a sequence labeling model which adapts BIO tagging scheme. The training process and the parameters settings of the BIO model are the same as SpanOIE  except that the output of the BiLSTM layer is directly fed into a feed forward network to do the classification for BIO labels. To focus on the model own strength, our model will not include dependency embeddings in the input representation unless stated otherwise. 

\begin{table*}[t!]
\begin{center}
\begin{tabular}{|l|l|}
\hline 
Orginal sentence & \emph{The keys, which were needed to access the building,}\\
&  \emph{were locked in the car.}\\ 
\hline 
SpanOIE & [\emph{The keys}; \emph{were needed}; \emph{to access the building}]\\
         & [\emph{The keys}; \emph{\textbf{to access}}; \emph{the building}]\\
         & [\emph{\textbf{The keys}}; \emph{were locked}; \emph{in the car}]\\
\hline
OpenIE4 & [\emph{The keys}; \emph{were needed}; \emph{to access the building}]\\
         & [\emph{The keys, which were needed}; \emph{were locked}; \emph{in the car}]\\
\hline
\end{tabular}
\end{center}
\caption{Example of extractions from Span OIE and OpenIE 4}
\label{examples}
\end{table*}

\subsection{Results on OIE2016}
We use OIE2016 to evaluate the precision and recall of different Open IE systems. The Precision-Recall (P-R) curve is shown in Figure~\ref{results_old}. The P-R curve is obtained by selecting a list of confidence score thresholds for each Open IE system. All the extractions below this threshold are dropped. We see that the SpanOIE system outperforms all other systems. Although our model is trained from the corpus bootstrapped from the extractions of OpenIE4, it still has a better performance than OpenIE4 as it is capable of using truly good extractions no matter how the threshold varies. The improvement of our model over OpenIE4 is two-fold. First, our model can find more predicates than OpenIE4 which leads to higher recall on more extractions. Second, our model is more accurate in finding the correct arguments. As an example shown in Table~\ref{examples}, our model finds one more predicate [\emph{to access}] and the correct A0 span [\emph{The keys}].

\begin{table*}[t!]
\begin{center}
\begin{tabular}{|l|rl|ll|}

\hline \textbf{Systems} & \multicolumn{2}{|c|}{\textbf{OIE2016}} & \multicolumn{2}{|c|}{\textbf{Re-OIE2016}} \\ \hline
& AUC & F1 &AUC &F1 \\
\hline
Stanford \cite{Angeli2015LeveragingLS} & 0.079 & 13.55& 0.115 & 16.70\\
OLLIE \cite{Mausam2012OpenLL} & 0.202 &38.58 & 0.313 & 49.47\\
PropS \cite{Stanovsky2016GettingMO} & 0.320 & 54.38& 0.433 & 64.16\\
ClausIE \cite{Corro2013ClausIECO} & 0.364 & 58.01& 0.464 & 64.17\\
OpenIE4 \cite{Mausam2016OpenIE} & 0.408 & 58.83& 0.509 & 68,32\\
\hline
BIO & 0.462 & 68.55 & 0.579 & 77.08\\
Seq2seq OIE \cite{Cui2018NeuralOI} & 0.473 & - & - & - \\
SpanOIE\_prev  &\textbf{0.491} & \textbf{69.42}& 0.647 & 78.08\\
SpanOIE & 0.489 & 68.65& \textbf{0.659} & \textbf{78.50}\\
\hline
\end{tabular}
\end{center}
\caption{\label{auc} AUC and the highest F1 score for different Open IE systems for OIE2016 and Re-OIE2016. SpanOIE\_prev is the model trained from the extractions with confidence score higher than 0.9 of OpenIE4. The rest settings are the same as SpanOIE.}
\end{table*}

\subsection{Results on Re-OIE2016}
We also evaluate the Open IE systems on Re-OIE2016 corpus. The P-R curves are shown in Figure~\ref{results_new}. We also compare the Area under P-R curve (AUC) and the best F1 score for different Open IE systems on two corpora in Table~\ref{auc}. Note that the code for the work of \cite{Cui2018NeuralOI} is not released and it is hard to reproduce their result. So we just report their results obtained from their paper. 

From the comparison, we see that although the performance ranking of Open IE systems is kept the same between OIE2016 and Re-OIE2016, the AUC value and the best F1 score increase for all the Open IE systems on Re-OIE2016. Comparing P-R curves, we see that the increase of AUC value mainly comes from the increase of recall. In fact, the number of extractions is fewer in Re-OIE2016 than OIE2016. The decrease of extractions mainly comes from three aspects that have been mentioned in previous sections. First, OIE2016 considers all the verbal adjectives as predicates which are deleted in Re-OIE2016. Second, it also excludes the extractions that take appositive as A0. 

Third, OIE2016 breaks conjunctions in arguments to generate multiple extractions. For example, given a sentence [\emph{Ben and Jenny went to Japan}], if the extracting is done on both sides of the conjunction, we have [\emph{Ben; went to; Japan}] and [\emph{Jenny; went to; Japan}] for extractions. However, since most existing Open IE systems (including ours) do not break the conjunctions, we preserve the conjunction constituent in an extraction instead of breaking them. Actually, the conjunction structures can be better postprocessed after the works of Open IE. Note that there are also numbers of wrong extractions in OIE2016 due to the incorrect break of conjunctions. Thus the extractions over the conjunctions are not taken into account in our Re-OIE2016 for evaluating Open IE systems.

\subsection{Usage of Training Corpus}
During the training, we make use of every extraction from OpenIE4 in the training corpus by integrating the confidence score into the loss function. Following \cite{Cui2018NeuralOI} which keeps only the extractions with confidence score higher than 0.9, we compare our span model to every Open IE system on both OIE2016 and Re-OIE2016. From Table~\ref{auc}, we can see that the performance of two approaches is very close, which indicates that the use of extractions with low confidence score does not hurt much the performance. 

We find that our method helps improve the model in some cases. For example, for the sentence [\emph{And he was in Ali 's army in the Battle of Jamal and later it was Muhammad ibn Abu Bakr who escorted Aisha back to Madina}] in OIE2016, for the predicate \emph{was}, the extraction from previous model is [\emph{Muhammad ibn Abu Bakr; was; in Ali 's army}] and the extraction from  our model is [\emph{he; was; in Ali 's army; in the Battle of Jamal}]. From this example, we see that our model is capable of performing pronoun argument extracting while the previous model cannot due to the absence of training corpus with pronoun argument.

Besides the extraction for pronoun arguments, we also observe that our model has improvements on other extraction patterns. For example, for the sentence [\emph{Yost was named manager of the Kansas City Royals , replacing Trey Hillman}] with the predicate \emph{replacing}, the extraction from the previous model is [\emph{the Kansas City Royals; replacing; Trey Hillman}] while the extraction from our model is [\emph{Yost; replacing; Trey Hillman}]. This is an extraction pattern where the predicate is in the form of present participle while the subject is far from the predicate. 

\subsection{The Impact of Syntax}
\begin{figure}[ht]
\begin{center}
\rule[-.5cm]{0cm}{0cm}\includegraphics[scale=0.5]{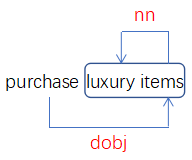}\rule[-.5cm]{0cm}{0cm}
\caption{Syntactic dependency tree for [\emph{luxury items}]}
\label{dp_span}
\end{center}
\end{figure}

To explore if syntactic clues may help Open IE, we introduce syntactic dependency labels into the word level representation of our model. The dependency parsing is performed by spaCy. Based on the span feature designed in Equation (\ref{span_feature}), we add another vector in this feature: 
\begin{equation}
    \label{span_syn}
    \begin{split}
        f_{span}(s_{i:j}) = h_i \oplus h_j \oplus h_i + h_j\\
         \oplus hi - h_j \oplus h_{dp(s_{i:j})}
    \end{split}
\end{equation}
where $dp(s_{i:j})$ is the index of the parent word of syntactic head of the span $s_{i:j}$. For example, Figure~\ref{dp_span} illustrates syntactic dependency parse tree for the span [$luxury$ $items$], which is a part of the sentence in Figure~\ref{model}. The syntactic head of this span is the word \emph{items}, whose parent is the word \emph{purchase}. So $h_4$, which is the BiLSTM output for the word \emph{purchase}, is also taken as a part of the span features in this example. Syntactic parse tree discloses the nonlinear structure for a span and is supposed to make a help in Open IE because all the words are not equally important in the same span. For example, in the span [\emph{luxury items}], the syntactic clue indicates that the word \emph{items} is more important than the word \emph{luxury} in finding the correct label for this span. The latter one might be more helpful in determining the boundary of the span.

Using the syntax-aware span feature as Equation (\ref{span_syn}), we re-train our model and achieve AUC 0.487 and 0.639 on OIE2016 and Re-OIE2016, respectively, which are both slightly lower than our original syntax-agnostic model. Such experimental results undoubtedly strike the belief on the helpfulness of syntactic information, which may be attributed to the noise inside the training corpus and imperfect predicted syntactic parsing trees. 

Despite the unfavorable performance, we still found that the span level syntactic information helps improve the extraction in some cases. For example, given a sentence [\emph{The dialects they speak are similar but have different intonations}], with the predicate \emph{have}, syntax-agnostic model gives the extraction [\emph{they; have; different intonations}] but the extraction from a syntax-aware model is [\emph{The dialects; have; different intonations}]. This example suggests that syntax is indeed helpful in Open IE, and it is hopefully proved that an effective syntax-aware Open IE system can be built on training corpus of higher quality and more accurate syntactic inputs.

\subsection{Limitations}
This subsection discusses cases on limitations of the current Open IE systems, including ours.

Open IE systems like OLLIE, OpenIE4 and ClausIE also extract context of a sentence while most of the latest neural model cannot do so
. For example, given a sentence [\emph{He believes The Lakers will win the game}], the correct extraction should be [\emph{context[He believes]; The Lakers; will win; the game}] because [\emph{The Lakers will win the game}] is not true according to the original sentence. That is also an important difference between Open IE and SRL since Open IE needs to identify correct and useful information. The reason why we did not bootstrap context information from OpenIE4 is that the context information extracted by OpenIE 4 is very noisy. Second, most of neural Open IE systems including ours cannot extract is appositive relation. For example, given a sentence [\emph{Obama, president of USA, gave a speech on Friday}], one extraction should be [\emph{Obama; be; president of USA}].

Third, we still need to explore better ways in constructing and leveraging training corpus for Open IE. One possible way is that we can bootstrap tuples from different existing Open IE systems instead of just from OpenIE4 so as to leverage the strength of different systems.

\section{Conclusions}
In this work, we bootstrap a large training corpus for Open IE and propose an improved model design to more sufficiently exploit the training data. We also present our accurately re-annotated benchmark test set (Re-OIE2016). At last, we introduce SpanOIE, which is the first span model for $n$-ary Open IE. Our model achieve new state-of-the-art performance on both benchmark evaluation datasets.

\bibliography{AAAI-2020}
\bibliographystyle{aaai}
\end{document}